# Model selection meets clinical semantics: Optimizing ICD-10-CM prediction via LLM-as-Judge evaluation, redundancy-aware sampling, and section-aware fine-tuning

Hong-Jie Dai[1,2,3,4], Zheng-Hao, Li[1], An-Tai Lu[5], Bo-Tsz Shain[1], Ming-Ta Li[5], Tatheer Hussain Mir[1], Kuang-Te Wang[6,7†], Min-I Su[6,7†], Pei-Kang Liu[5,8,9†], Ming-Ju Tsai[10,11*†]

[1]Intelligent System Lab, College of Electrical Engineering and Computer Science, Department of Electrical Engineering, National Kaohsiung University of Science and Technology, Kaohsiung, Taiwan

[2]National Institute of Cancer Research, National Health Research Institutes, Tainan, Taiwan

[3]Center for Big Data Research, Kaohsiung Medical University, Kaohsiung, Taiwan

[4]School of Post-Baccalaureate Medicine, Kaohsiung Medical University, Kaohsiung, Taiwan

[5]Department of Healthcare Administration and Medical Informatics, Kaohsiung Medical University, Kaohsiung, Taiwan

[6]Division of Cardiology, Department of Internal Medicine, Taitung MacKay Memorial Hospital, Taitung, Taiwan

[7]Department of Medicine, MacKay Medical University, New Taipei, Taiwan

[8]Department of Ophthalmology, Kaohsiung Medical University Hospital, Kaohsiung Medical University, Kaohsiung, Taiwan.

[9]School of Medicine, College of Medicine, Kaohsiung Medical University, Kaohsiung, Taiwan

[10]Division of Pulmonary and Critical Care Medicine, Department of Internal Medicine, Kaohsiung Medical University Hospital, Kaohsiung Medical University, Kaohsiung, Taiwan

[11]Department of Internal Medicine, School of Medicine, College of Medicine, Kaohsiung Medical University, Kaohsiung, Taiwan

## Abstract

Accurate International Classification of Diseases (ICD) coding is critical for clinical documentation, billing, and healthcare analytics, yet it remains a labour-intensive and error-prone task. Although large language models (LLMs) show promise in automating ICD coding, their challenges in base model selection, input contextualization, and training data redundancy limit their effectiveness.

We propose a modular framework for ICD-10 Clinical Modification (ICD-10-CM) code prediction that addresses these challenges through principled model selection, redundancy-aware data sampling, and structured input design. The framework integrates an LLM-as-judge evaluation protocol with Plackett-Luce aggregation to assess and rank open-source LLMs based on their intrinsic comprehension of ICD-10-CM code definitions. We introduced embedding-based similarity measures, a redundancy-aware sampling strategy to remove semantically duplicated discharge summaries. We leverage structured discharge summaries from Taiwanese hospitals to evaluate contextual effects and examine section-wise content inclusion under universal and section-specific modelling paradigms. Experiments across two institutional datasets demonstrate that the selected base model after fine-tuning consistently outperforms baseline LLMs in internal and external evaluations. Incorporating more clinical sections consistently improves prediction performance. This study uses open-source LLMs to establish a practical and principled approach to ICD-10-CM code prediction. The proposed framework provides a scalable, institution-ready solution for real-world deployment of automated medical coding systems by combining informed model selection, efficient data refinement, and context-aware prompting.

* Corresponding author at: Kaohsiung Medical University Chung-Ho Memorial Hospital, Kaohsiung, Taiwan Republic of China. E-mail addresses: siegfried.tsai@gmail.com.

† These authors contributed equally to this work.

## 1. Introduction

The International Classification of Diseases (ICD), established by the World Health Organization (WHO) in 1948 [1], provides a global standard for recording, reporting and monitoring diseases across international health systems. Many countries have adopted localized adaptations of ICD to meet national clinical and administrative needs. For instance, the United States utilizes ICD-10-CM (Clinical Modification) and ICD-10-PCS (Procedure Coding System) for diagnosis and procedural coding. In Taiwan, the Ministry of Health and Welfare at the National Health Insurance Administration mandated the transition from ICD-9-CM to ICD-10-CM/PCS in 2016, adopting the 2014 ICD-10-CM/PCS standard that comprises approximately 71,900 diagnosis codes and 78,500 procedure codes.

Manual ICD-10 coding is a labor-intensive, error-prone and time-consuming process requiring specialized expertise to interpret complex medical records and assign appropriate codes. A recent report from Linkou Chang Gung Memorial Hospital estimated that certified coding specialists (CCSs) spend an average of 22 minutes per record for manual coding [2], equating to over 27 full-time working days per coder each month. This inefficiency, coupled with increasing documentation volume, underscores the need for scalable, accurate, and automated ICD-10 coding solutions.

Large language models (LLMs) have emerged as promising tools for automating various tasks related to natural language understanding. However, their application to ICD coding remains unique challenge as accurate ICD coding requires more than linguistic fluency, it demands deep clinical reasoning, contextual inference, and precise mapping to hierarchical medical terminologies. Prior studies [3-5] have shown that even state-of-the-art proprietary models (e.g., GPT-4, Claude, Gemini) and open-source LLMs (e.g., LLaMA, Mistral), including those pretrained on biomedical corpora [4], struggle to match the reliability and precision of CCS without further domain-specific tuning. Issues such as insufficient domain grounding, hallucination, and lack of alignment with clinical curation practices compared to CCSs are common limitations [5]. Although prior work has shown that fine-tuning improves performance [6-9], most of the previous experiments were not conducted on real-world curation data, and selecting a well-suited base model and aligning it with clinical documentation practices remains a critical challenge [10].

Meanwhile, the adoption of HL7/FHIR standards has accelerated the structured documentation of discharge summaries worldwide. Taiwan's electronic medical record (EMR) formats [11], which include structured clinical sections, such as discharge diagnosis (DischgDiag), medical history (MedHist) and operation note (OpNote), enable systematic investigation of how section-wise content impacts coding accuracy. Additionally, redundancy is a known characteristic of clinical notes [12]. For example, in our dataset, we observed that certain frequently occurring ICD-10 codes, such as I10 (essential hypertension), Z51.11 (encounter for antineoplastic chemotherapy), and E11.9 (type 2 diabetes mellitus without complications), frequently appear in similar narrative contexts, offering diminishing returns in model training.

Motivated by these insights, we propose a modular and extensible pipeline for ICD-10-CM coding using open-source LLMs, incorporating principled model selection, redundancy reduction, and context-aware prompting. The core contributions of this study are as follows:

- Base model selection via LLM-as-judge [13]: We propose a pairwise comparison framework to assess the intrinsic ICD-10 comprehension among candidate models. Matchup results are aggregated using the Plackett-Luce model [14, 15] to derive interpretable model rankings and guide base model selection.
- Redundancy-aware sampling: Inspired by our previous word embedding-based synthetic minority over-sampling technique [16], we implement an embedding-based deduplication strategy that identifies semantically redundant discharge summaries, improving training efficiency and model generalizability without sacrificing code diversity.
- Section-aware prompting: Leveraging Taiwan's structured EMR format [11], we investigate the impact of incorporating multiple clinical sections on coding performance under both universal and section-specific modelling strategies.
- Robustness evaluation across institutions: We validate the generalizability of our pipeline using datasets from two geographically and administratively distinct hospitals, accessing both intrinsic model comprehension and downstream code prediction accuracy without retraining.

## 2. Related works

The task of automated medical code prediction has received increasing attention in recent years, especially with the widespread adoption of ICD-10-CM/PCS across healthcare systems. Traditional machine learning approaches, such as naïve Bayes [17] and support vector machines [18], have gradually replaced deep learning models that leverage clinical text representations for multi-label prediction. Hierarchical attention networks (HAN) [19] and recurrent neural architectures such as BiGRU [20] have demonstrated success, particularly when applied to publicly available benchmark datasets like MIMIC-III [21]. However, most of these datasets are limited to ICD-9 codes and lack consistent section-level structure, thereby restricting their utility for exploring context-aware modelling strategies.

Recent efforts have shifted toward LLMs, which offer the potential to perform ICD code prediction directly from free-text narratives with minimal task-specific supervision. Studies such as Ji, et al. [8] and Suvirat, et al. [22] fine-tuned BERT for ICD coding and reported notable performance gains compared to traditional methods. Subsequent works by Dai, et al. [7] and Nawab, et al. [6] explored decoder-based models, such as GPT-2 and GPT-3.5 turbo, showing that autoregressive LLMs could outperform encoder-based counterparts for medical code generation when paired with instruction-style prompts.

Despite their promise, general-purpose LLMs often underperform in medical coding scenarios due to insufficient domain alignment and weak grounding in structured medical terminologies. Multiple studies [3-5] have shown that even proprietary models fail to achieve competitive results in medical code generation when prompted directly, even under expert-designed prompt conditions. Boyle, et al. [23] demonstrated that incorporating ICD diagnostic ontology into prompting does not fully mitigate the lack of task-specific fine-tuning: LLaMA-2, GPT-3.5, and GPT-4 all underperformed compared to BERT-based models in structured ICD-10 code retrieval tasks. Similarly, Puts, et al. [24] emphasized the need of aligning model behavior with the actual workflow of clinical coders, including inferring codes through multi-step reasoning, cross-referencing across sections, and excluding irrelevant diagnoses, rather than relying solely on surface-level narrative matches. Additionally, hallucination and instruction-following failures remain prevalent in general-purpose models not fine-tuned for medical coding workflows, limiting their reliability in real-world applications [5].

These challenges highlight the continued importance of domain-specific fine-tuning. However, a critical bottleneck lies in base model selection: the choice of an initial model heavily influences fine-tuning success. A poor base model can result in negative transfer, as seen in transfer learning for computer vision [25]. In the context of ICD coding, Lee and Lindsey [4] systematically evaluated a suite of off-the-shelf open-source LLMs, including Llama2 [26], Meditron [27], MedLlama2 and Mistral [28], to assess their intrinsic understanding of ICD codes via definition matching. Their findings revealed that even the best-performing open-source model, Mistral, and proprietary model, GPT-4, failed to capture the semantic structure of ICD definitions fully, underlining the difficulty of this task and the limitations of current LLMs for downstream ICD applications.

Traditionally, the most accurate method for base model selection requires fully fine-tuning each model and comparing downstream performance, a costly and often impractical process, especially when exploring large repositories such as Hugging Face. To mitigate this, Bai, et al. [10] proposed a model selection method for computer vision by mapping pretrained models and historical tasks into a joint transfer-related latent subspace, where the distance between model vectors and task vectors represents their transferability. A large vision model is then used as a proxy to infer the vector for a new task in the transfer-related space, enabling inner-product-based transfer score estimation. However, this approach is tailored to vision models, with task structures, embeddings and experimental configurations that are not directly applicable to the language modelling domain.

To address the model selection problem in LLM for ICD coding, we propose a principled based model selection framework grounded in LLM-as-a-judge paradigms [13]. Our approach evaluated candidate models based on their intrinsic comprehension of ICD-10-CM definitions via pairwise comparison of their generated code descriptions. The match outcomes are aggregated using the Plackett-Luce model [14, 15] to derive a robust and interpretable ranking for base model selection. This framework enables computationally efficient filtering of candidate models before fine-tuning, serving as a practical proxy for downstream performance.

Moreover, while many benchmark datasets such as MIMIC-III [21] lack section-level structure and are confined to ICD-9 codes, we compiled a semi-structured ICD-10-CM coding dataset from a Taiwanese medical Centre. This dataset aligns with Taiwan's EMR template [11], containing structured clinical sections. This format enables systematic investigation of how different combinations of clinical section contents impact coding performance.

Finally, clinical notes often exhibit semantic redundancy [12]. Previous studies have demonstrated that deduplication preserves model performance, reduces training time, decreases dataset size, and even enhances predictive accuracy [29, 30]. Inspired by these findings, we propose a redundancy-aware sampling strategy that filters semantically similar records based on embedding similarity and code overlap, thereby reducing training cost without compromising model accuracy.

## 3. Methods

Fig. 1 presents an overview of the proposed modular pipeline for developing an ICD-10-CM coding system, highlighting its structured methodology in contrast to conventional approaches based on ad hoc or intuition-driven model selection. The left portion of the figure illustrates the proposed framework, which begins with data acquisition, including identifying the top-50 most frequently occurring ICD-10-CM codes in the collected dataset and candidate LLM acquisition. A preprocessing stage follows, incorporating a redundancy-aware sampling strategy to remove semantically repetitive discharge summaries. Next, we evaluated a set of candidate LLMs through pairwise comparisons using the LLM-as-judge strategy, which accesses each model's intrinsic understanding of ICD-10 code definitions. We then aggregate these results and rank them using the Plackett-Luce model, enabling principled selection of the most suitable base model. The top-ranked base model is subsequently selected and fine-tuned using section-aware instruction-style prompts derived from structured discharge summary content. The final fine-tuned model is deployed in a hospital setting to assist CCSs in real-world coding workflows environment to support automated medical coding.

The modular pipeline contrasts traditional "gut-feeling"-based approaches, which lack principled guidance and offer no optimality guarantees. Such ad hoc methods often result in inefficient resource usage, particularly in the base model selection phase, where all candidate LLMs must be fine-tuned and evaluated on the full dataset to identify the best-performing model. This process is both time-consuming and computationally expensive. We then elaborate on the components of the proposed framework in the following subsections.

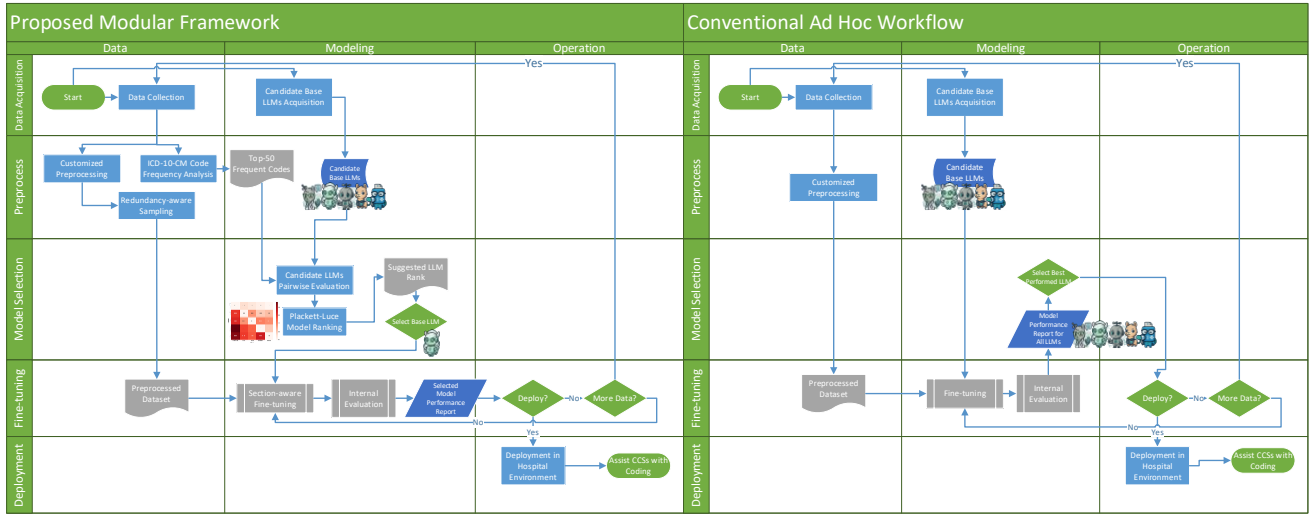

**Fig. 1.** Overview of the proposed modular framework for developing an ICD-10-CM coding system. The left portion depicts the structured pipeline for model development, including redundancy-aware preprocessing, LLM-as-judge along with Plackett-Luce model selection, section-aware fine-tuning, and deployment. The right portion contrasts this approach with conventional, intuition-driven pipelines lacking principled model selection and optimization.

3.1. Dataset

We used two datasets from distinct hospitals in Taiwan. The first internal dataset supports the development and deployment phases of the developed system. This dataset comprises anonymized discharge summaries from Kaohsiung Medical University Chung-Ho Memorial Hospital (KMUH) in Kaohsiung, Taiwan, from April 2019 to March 2021. In addition, a separate set of discharge summaries was collected between October and December 2024, after the deployment of the prototype system into the routine work of CCS, as detailed in our previous work[7]. The second dataset serves as another external validation set to evaluate the robustness and generalizability of the developed model in cross-institutional settings without retraining. This dataset was collected from Taitung MacKay Memorial Hospital (TMMH) in Taitung, Taiwan and includes discharge summaries recorded between January 2021 and December 2022.

In contrast to widely used datasets such as MIMIC-III[21], which exclusively contain ICD-9 codes, we annotated our datasets fully with ICD-10 codes. Furthermore, the discharge summaries adhere to structural conventions defined by the Taiwan electronic medical record template [11], which organizes clinical content in a semi-structured, section-based format. Table 1 illustrates how we divide each collected discharge summary into five major sections, with example content provided for each. The textual content within each section is expressed in deidentified natural language, maintaining the narrative style commonly found in clinical documentation. Following consultation with CCSs, we confirmed that the DischgDiag section serves as the primary basis for ICD coding. The final column of Table 1 presents the section-checking priority rankings recommended by CCSs, reflecting the relative importance of each section when reviewing discharge summaries for ICD-10-CM and ICD-10-PCS code assignment.

**Table 1**

A deidentified example of a semi-structured discharge summary used in this work, following the section-based format.

| Section Name | Section Text | Suggested Priority |
| --- | --- | --- |
| Discharge Diagnosis | 1. Right distal radius fracture<br> - post open reduction and internal fixation of right distal radius fracture on 2043/6/2<br>--- Underlying diseases ---<br>. Left hydronephrosis, status post double-J insertion on 2041/09/25, 2042/7/14<br>. Recurrent cervical cancer with metastatic left iliopoas muscle adenocarcinoma … | 1 |
| Medical History | ▋HPI (History of Present Illness):<br>Patient was a 67-year-old woman with underlying disease of:<br>. Left hydronephrosis status post double-J insertion on 2031/09/25, 2042/7/14<br>. Recurrent cervical cancer with metastatic left iliopoas muscle adenocarcinoma<br>…<br>She fell down on 5/30 in morning resulting in right wrist painful disability. She was brought to our emergent room and then …<br>▋PH (Past History): + Past History<br>DM (diabetes mellitus): denied<br>Hypertension: +<br>CKD (chronic kidney disease): denied | 3 |
| Pathology Report | no exam during admission. | 4 |
| Operation Note | Open reduction and internal fixation of right distal radius fracture on 2043/6/2 | 2 |
| Treatment Course | After admission, this 73-year-old female received our general survey. She underwent Open reduction and internal fixation of right forearm smoothly on 2043/6/2, and no peri-operative complications were noted. She had right forearm operation pain (Visual Analogue Scale:3/10). We gave her Cefazolin and … | 5 |

3.2. Preprocessing

Although the collected dataset was curated by CCSs and annotated with ICD-10 codes, we observed that some reports either lacked corresponding codes or included incomplete or erroneous entries (*e.g.*, invalid or non-standard ICD codes). We implemented a multi-step data cleaning and deduplication pipeline to address these issues.

*3.2.1 Data cleaning procedure*

Firstly, we excluded the discharge summaries that had missing ICD-10 code assignments or invalid codes from the dataset. We then applied character-level normalization, standardizing punctuation marks across Chinese and English texts and removing extraneous whitespaces and line breaks. We applied rule-based pattern matching to strip non-clinical segments, such as administrative notes and timestamped metadata, to retain only clinically relevant content. After cleaning, the dataset was partitioned into training, development and test sets using multi-label stratified sampling [31], preserving the label distribution with an 8:1:1 split ratio.

*3.2.2 Redundancy-aware sampling strategy*

To further enhance data quality and training efficiency, we employed a semantic similarity-based deduplication method using sentence embeddings and approximate nearest neighbour search [32]. This approach targeted near-duplicate summaries that shared identical ICD-10-CM codes, which offers limited additional learning values. Specifically, first, we encoded the discharge summary into a semantic vector using the pretrained all-MiniLM-L6-v2 sentence transformer [33]. We then constructed an FAISS index [34] to facilitate fast vector search, and computed the L2 distances between embeddings as proxies for semantic similarity. For each summary, we identified its nearest semantic neighbour. If both records shared identical ICD-10 codes and their semantic similarity exceeded a threshold (empirically set to 0.9 based on preliminary observations of the semantic similarity distribution within the dataset), we marked them as redundant and subject to removal.

To determine which summary to retain from each redundant pair, we compute the perplexity (PPL) of both candidates using the selected base LLM. One summary exhibited a PPL at least 5% higher than its counterpart, indicating greater linguistic uncertainty and potentially more diverse language, which we retained. If PPLs were within the threshold, we preserved the longer summary to retain more contextual information.

3.3. Pretrained base model selection

Given the known limitations of general-purpose LLMs in understanding and extracting ICD-10-CM codes [3-5], selecting an appropriate pretrained base model is a critical step in developing a robust and reliable medical coding system. Rather than relying solely on superficial factors such as model size or popularity, we adopt an evaluation strategy that prioritizes a model's intrinsic ability to comprehend the semantic meaning of ICD-10 codes. Specifically, we formulate the selection process as a pairwise preference task, in which we compare the outputs of two candidate LLMs to determine which model demonstrates better understanding. For each comparison, we prompt both models to generate the medical conditions associated with a given ICD-10-CM code, as illustrated in Table 2. We then evaluate these responses to assess semantic relevance and clinical accuracy, as a proxy for the model's latent comprehension of medical coding concepts.

**Table 2**

Prompts used for two candidate LLMs and the judge LLM.

Instead of manually comparing the textual outputs of candidate LLMs to reference answers [30], counting the number of words overlaps with gold answers [4], or relying on similarity metrics such as cosine similarity [3], we adopted the LLM-as-judge strategy [13] to directly evaluate each model's intrinsic understanding of ICD-10 codes. Specifically, we employed a lightweight language model, Atla Selene Mini [35], to perform pairwise assessments of candidate models' comprehension of ICD-10 semantics through a series of prompt-based probes. We constructed these probes using the top 50 most frequently occurring ICD-10-CM codes in our training corpus. This approach enables a systematic, scalable, and model-agnostic evaluation framework, providing an evidence-driven basis for selecting the most appropriate base model for downstream ICD-10 coding fine-tuning.

Table 1 is the prompt template used by the judge model. It is a modified version of the original template proposed by Alexandru, et al. [35], tailored to capture better clinical specificity and terminology of ICD-10 coding tasks. While the judge model typically adhered to the expected format using "A" or "B" to indicate a preferred candidate, we observed occasional deviations, such as outputs like "2 (Note: This result does NOT follow your format)". To ensure robustness and consistency in the systematic evaluation, we implemented multiple regular expressions to accommodate these variants and extract the final decision. In cases where the judge LLM's response does not clearly indicate a preference, we record the comparison as a tie.

After completing all pairwise comparisons among the $k$ candidate models, we construct a directed comparison graph $G = (V, E)$, where $V = \{1, ..., k\}$ and an edge $(j, i) \in E$ indicates that model $i$ wins at least once against model $j$ in the matchup set $D$. Let $D = \{(c_\ell, A_\ell) | \ell = 1, ..., d\}$ denote a collect of $d$ independent matchup observations, where $c_\ell$ represents the selected (winning) model and $A_\ell$ the set of competing alternatives in that matchup. Following Luce [14]'s choice axiom, we define the probability of selecting a base model $i$ from an alternative opponent set $A_i$ as $p(m_i|A_i)$. For any two models $m_i$ and $m_j$ within alternative sets $A_i$ and $A_j$, respectively, the axiom asserts that the odds of selecting model $m_i$ over model $m_j$ are independent of the rest of the alternative matchups, as shown in (1).

$$\frac{p(m_i|A_i)}{p(m_j|A_i)} = \frac{p(m_i|A_j)}{p(m_j|A_j)} \qquad (1)$$

This leads to the Plackett-Luce mode, which generalizes $k$-way ranking tasks and can be solved by finding the transition matrix of a Markov chain given its stationary distribution [36]. In our base model selection task, we treat the win-rate matrix as the transition matrix $\lambda$ and determine the optimal base model by estimating the stationary distribution $\boldsymbol{\pi} = [\pi_i]$. The global balance equations that relate the transition rates $\lambda$ to the stationary distribution $\boldsymbol{\pi}$ are defined as follows:

$$\sum_{j \neq i} \pi_i \lambda_{ij} = \sum_{j \neq i} \pi_j \lambda_{ji} \qquad \forall i \qquad (2)$$

The stationary distribution is therefore invariant under rescaling of the transition rates. Under this formulation, the probability of selecting model $i$ from a set of alternatives $A_i$ is given by (3).

$$p(m_i|A_i) = \frac{\pi_i}{\sum_{j \in A_i} \pi_j} \tag{3}$$

This implies that the selection preference is parameterized by $\pi_i$, where the normalization constraint $\sum_i \pi_i = 1$ holds.

To estimate the distribution, we employed the iterative Luce spectral ranking (ILSR) algorithm developed by Maystre and Grossglauser [37]. The resulting stationary distribution $\pi$ serve as our final ranking criterion for base model selection. The model with the highest stationary probability $\pi_i$ is selected as the optimal base model for downstream ICD-10-CM coding tasks.

### 3.4. Section-wise instruction turning

After selecting the base model, we fine-tuned it using the training dataset. Based on our previous experience with decoder-only architectures for ICD-10 coding, as well as findings by Yoo and Kim [9], we adopted a standard text generation loss for training. Empirically, this approach has proven more effective than more complex strategies such as label attention mechanism [38] or knowledge injection techniques [9]. In the proposed implementation, the base model is trained to generate a target output sequence $\mathcal{Y} = [y_1, y_2, \ldots, y_m]$, which represents the set of ground-truth ICD-10-CM codes assigned to a discharge summary. This output is conditioned on an input sequence $X = [x_1, x_2, \ldots, x_n]$ corresponding to the section-wise content of the summary.

We designed an instruction-style section-wise prompt template (illustrated in Fig. 2) to guide supervised fine-tuning. Followed the instruction design principle outlined by Bsharat, et al. [39], each section of the discharge summary is explicitly introduced with the prefix "###", followed by the section name and its corresponding content. This formatting strategy reinforces semantic segmentation within the clinical text, helping the model to better distinguish between different components of the summary. The model is fine-tuned to maximize the conditional likelihood of the target output sequence $\mathcal{Y}$ given the input $X$, using the standard autoregressive language modeling objective:

$$P(\mathcal{Y}|X) = \prod_{i=1}^{m} p(y_i|X, y_{<i}) \tag{4}$$

The output format is explicitly structured to differentiate between the primary code and other codes using the format:

"MAINCODE: {main_code}\nOTHERCODE: {other_codes}".

This output convention aligns with clinical coding practices by encouraging the model to prioritize the assignment of the principal diagnosis from other associated conditions.

**Fig. 2.** Instruction tuning template with section-wise information for the ICD-10-CM coding task.

To evaluate the impact of section-wise content inclusion, we develop a universal model and several section-combination-specific fine-tuned models based on the section priority rankings suggested by CCSs, as shown in the last column of Table 1. We trained the universal models using the full template depicted in Fig. 2, in which missing section contents were explicitly filled with "Nil" during training and inference. This design allows the model to flexibly process any subset of the defined sections, depending on the actual structure of the input summary. In contrast, each section-combination-specific model was fine-tuned using the same prompt template but includes only the specific sections relevant to its designated configuration. All other sections, including their headers, were entirely omitted to reflect the intended structural scope. When required sections were missing from an input summary, we removed the corresponding section headers during training and inference.

3.5. Experimental setup

*3.5.1 Experimented LLMs and training details*

As stated in the Section 2, Lee and Lindsey [4] demonstrated that among several open-sourced LLMs, including Llama2 [26], Meditron [27], MedLlama2 and Mistral [28], Mistral achieved the best results in generating ICD-10 code definitions. Building on their findings, we focused on decoder-only architecture to evaluate the effectiveness of our proposed model selection strategy. We selected five candidate base models: PubMedGPT2, which was utilized in our previous work [7], Llama2-7B, Mistral-7B instruct, along with their respective domain-adapted variants MedLlama2 and BioMistral-7B [40].

For fine-tuning these models on downstream ICD-10 coding tasks, we adopted the quantized low-rank adaptation (QLoRA) technique [41] and 4-bit normal float quantization to accommodate hardware constraints. All performance were conducted using a single NVIDIA RTX A600 Ada GPU, with the input sequence length capped at 2,048 tokens. The models were trained using the AdamW optimizer with a learning rate of 5e-5.

*3.5.2 Evaluation metrics*

The primary evaluation metrics used in our ICD-10 coding experiments including micro-precision (P), micro-recall (R), and micro-F1-measure (F), which are commonly adopted for multi-label classification tasks. We report these metrics across two evaluation scopes: 1) the top-50 most frequent ICD-10 codes and 2) the full set of ICD-10 codes.

In addition, because the main diagnosis code plays a critical role in determining diagnosis related groups (DRGs) [7], we report an additional metric: main diagnosis code accuracy (MDCA), which is defined as follows: for each test case, if the model's predicted main diagnosis code exactly matches the ground truth main code, the prediction is considered correct. The MDCA is then calculated as the proportion of correct main code predications averaged across all test cases.

*3.5.3 Internal and external experiment designs*

We designed four experiments to evaluate the effectiveness of the proposed strategy. Experiment 1 focused on assessing the intrinsic ICD-10 comprehension of each candidate base model using the LLM-as-judge, along with ILSR described in Section 3.3. In this experiment, we prompted each model to generate definitions for the 50 most frequently occurring ICD-10-CM codes in our training corpus. The judge model then performed pair-wise comparisons between model outputs to generate a win-rate transition matrix, which was input for ILSR to identify the most suitable base model. In addition to reporting the final model selection, we visualize the win-rate matrix to facilitate interpretability and comparison across candidate models.

Experiment 2 evaluated the downstream ICD-10 coding performance of all candidate models. The experiment had two primary objectives: (1) to compare the performance of decoder-based models after instruction tuning against the model ranking inferred by ILSR, and (2) to conduct a comprehensive comparison of all developed models on the ICD-10 coding task. The evaluation was conducted on a held-out test set sampled from the dataset compiled at KMUH between 2019 and 2021 (see Section 3.1). Due to the token length constraints in certain models, such as PubMedGPT-2, which accepts a maximum of 1024 tokens, the input content had to be restricted. Consequently, we only used the DischgDiag section for the inputs in training and evaluation. To maintain consistency, we fine-tuned all decoder-based models in this experiment using the same prompt template containing only the DischgDiag section. We then compare the performance of the fine-tuned models against three baseline systems: 1) a BiGRU with BERT-based word embeddings proposed by Chen, et al. [20]; 2) a BERT-based model proposed by Pascual, et al. [42]; and 3) the hierarchical attention network (HAN) used by Dong, et al. [19].

Experiment 3 examined the effectiveness of the proposed redundancy-aware sampling strategy during training, using the best-performing base model identified in Experiment 2. Building upon that foundation, we further investigated the impact of incrementally incorporating section-wise content into the instruction prompt (see Fig. 2). Additional sections were introduced one at a time based on the priority rankings provided by CCSs, as listed in the second column of Table 3, to construct a series of section-combination-specific models. In parallel, a universal model was developed by including all defined sections simultaneously as described in Section 3.4. We assessed the model performance on the full test set and a subset of test samples containing all available section combinations to evaluate the influence of section inclusion. This experimental setup enables a controlled comparison of how different combinations of section-level information affect prediction accuracy for ICD-10-CM coding tasks.

Experiment 4 aimed to evaluate the robustness and generalizability of the developed models when applied to real-world, out-of-distribution hospital data. To this end, we conducted external validation using two independently curated datasets of structural discharge summaries collected from KMUH and TMMH. The KMUH dataset consists of summaries between October and December 2024. Eleven CCSs at KMUH curated these records in their routine clinical coding workflow. On the other hand, the external TMMH dataset comprises retrospective summaries collected from January 2021 to December 2022. Three CCSs at TMMU curated the data and reflected coding practices in a different institutional setting. To assess model robustness and real-world applicability, we applied all previously developed models to external datasets and evaluated their ICD-10-CM coding performance using standard metrics. This experiment offers critical insights into the extent to which models trained on data from a single institution can generalise to other hospitals with potentially different clinical documentation styles, coding behaviours, and temporal distributions.

## 4. Results and discussions

### 4.1. Dataset statistical information

#### 4.1.1 ICD-10-CM codes distribution analysis

The KMUH dataset, comprising discharge summaries collected between April 2019 and March 2021, served as the primary internal dataset for model training and testing. Following the preprocessing procedures described in Section 3.2, including removing discharge summaries with invalid or missing ICD-10 code assignments, 125,820 discharge summaries were retained with 11,991 unique ICD-10-CM codes. We partitioned the dataset using an 8:1:1 stratified split to support robust model development and evaluation, resulting in 100,656 training samples, 12,582 validation samples, and 12,582 test samples. Fig. 3 illustrates the distribution of ICD-10-CM codes by chapter across the three sets.

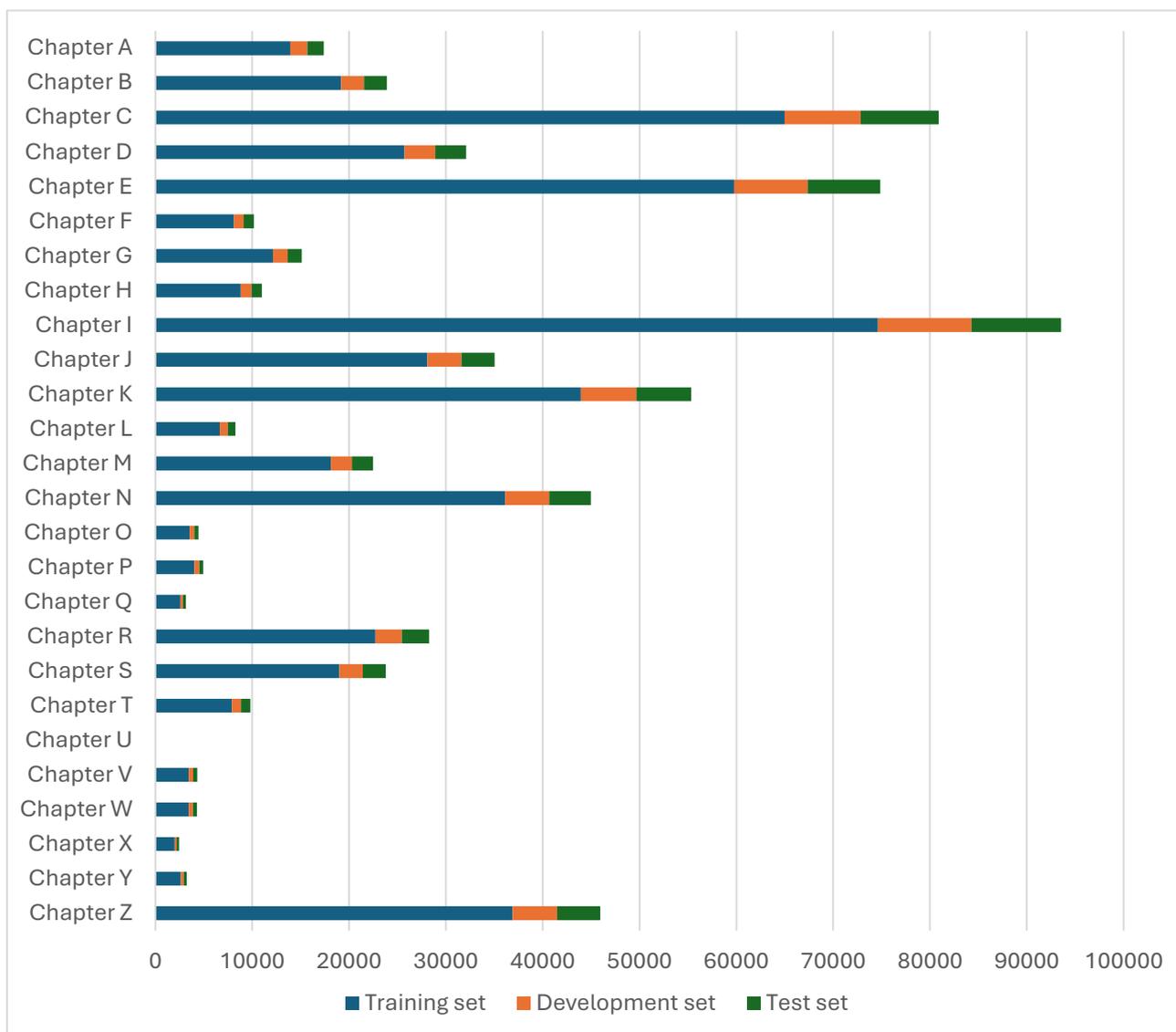

**Fig. 3.** Distribution of ICD-10-CM chapters in the training, validation, and test sets.

As shown in Fig. 3, certain ICD-10-CM chapters appear disproportionately high frequency, particularly chapters I, C, and E, which correspond to diseases of circulatory system diseases, neoplasms, endocrine, nutritional and metabolic diseases, respectively. In contrast, chapters such as X (diseases of the respiratory system), Q (congenital malformations, deformations and chromosomal abnormalities), and Y (external causes of morbidity and mortality) are underrepresented, while chapter U (codes for special purposes such as emergency codes additions by WHO) is absent.

*4.1.2 Section combination analysis and token length limitation estimation*

To better understand content heterogeneity and guide section-wise modelling, we further analysed the distribution of section combinations in the dataset. Since the DischgDiag section is the primary basis for ICD-10 code assignment, it was the anchor for all combination analysis. As shown in Table 3, the DischgDiag section is present in all 125,820 reports, underscoring its central role in discharge summary documentation. The MedHist section is the most frequently co-occurring section, with 124,624 reports containing both DischgDiag and MedHist. Of these, 35 contain only these two sections. Other common combinations involve the operation note (OpNote) followed by the pathology report (PathRep) and the treatment course (TreatCous). Notably, the combination comprising all five sections appears in more than half of reports (65,459), suggesting a standardised documentation structure adopted in a substantial proportion of the dataset.

**Table 3**

Distribution of section combinations in the KMUH dataset. Each row represents a unique combination of discharge summary sections in the dataset. The second column indicates the number of summaries containing the combination and those exclusively containing that specific combination. Based on tokenised input, the third column reports the percentage of summaries within each combination that exceed 2,048 tokens in length.

| Section Combination | Reports Containing Combination (Test) / Report Exclusively Containing Combination (Test) | Reports Containing Combination Exceeding 2,048 tokens (%) |
|---|---|---|
| (1) Discharge Diagnosis | 125,820 (12,582) / 0 (0) | 0 |
| (2) 1 + Medical History | 124,624 (12,463) / 35 (4) | 12.89 |
| (3) 1 + Operation Note | 86,376 (8,638) / 0 (0) | 12.60 |
| (4) 3 + Medical History | 86,341 (8,635) / 6 (1) | 12.60 |
| (5) 4 + Pathology Report | 65,514 (6,552) / 55 (6) | 14.04 |
| (6) 5 + Treatment Course | 65,459 (6,546) / 65,459 (6,546) | 14.03 |

We analysed the token length distribution associated with each section combination to assess the feasibility of processing the full contents for model training. The rightmost column in Table 3 presents the percentage of samples in each combination that exceed the 2,048-token threshold. For instance, 12.89% of summaries in the DischgDiag+MedHist group exceed the limit, compared to 14.03% in the full five-section group. These statistics suggest that setting the maximum input length at 2,048 tokens allows full coverage for approximately 85% of training samples. Due to hardware limitations, we adopted the 2,048-token limit for all downstream experiments. For samples exceeding this limit, we applied a priority-based truncation strategy as detailed in Table 1, where we first truncate lower-priority sections to preserve diagnostically critical content.

*4.1.3 Redundancy-aware sampling outcome*

AApplying the redundancy-aware sampling procedure described in Section 3.2, we reduced the training set from 100,656 to 85,820 summaries. Fig. 4 illustrates the ICD-10-CM chapter-level distribution before and after deduplication. Nearly all chapters experienced reduced sample counts, with chapter C exhibiting the most substantial decline. This pattern reflects the high degree of semantic redundancy within frequently occurring codes for inpatient samples, such as those associated with chronic or oncologic conditions.

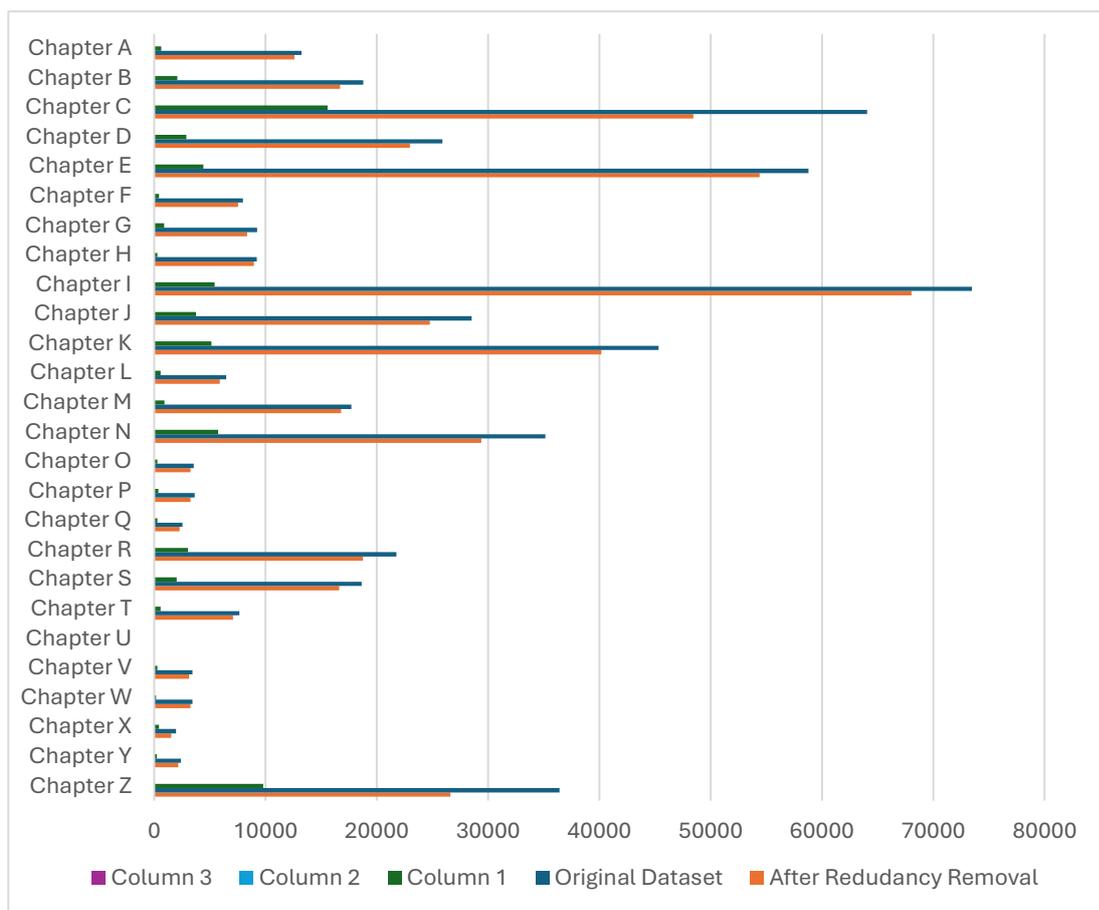

**Fig. 4.** ICD-10-CM chapter distribution (A-Z) in the training set before and after applying the redundancy removal procedure.

A Pearson correlation analysis yielded a strong positive correlation (r=0.808) between the original chapter-wise sample sizes and the corresponding reductions, indicating that the deduplication strategy effectively targeted overrepresented categories without indiscriminately pruning data. Importantly, the sampling procedure adequately represented less common chapters, preserving the distributional balance of clinically diverse diagnostic codes. These results confirm the effectiveness of the redundancy-aware sampling approach in improving data efficiency. By reducing training redundancy while preserving semantic diversity, the curated dataset better supports model generalization and enhances computational efficiency for downstream fine-tuning.

4.2. Experiment 1: Effectiveness of base model selection for ICD-10-CM coding task

### 4.2.1 Results of base model selection

Fig. 5 presents the results of the LLM-as-judge evaluation, which accesses the intrinsic ICD-10 comprehension capabilities of five base models through pairwise comparisons. The left panel visualizes the win-loss relationships among models using a mixed directed graph, where each node denotes a candidate base model. A directed edge from model A to model B indicates that A outperforms B in the ICD-10 code definition generation task. In cases where no clear winner is identified (*i.e.*, a tie), an undirected edge connects the models. The outdegree of each node reflects the number of wins achieved by the corresponding model. The right panel provides a detailed win-rate matrix heatmap, where each cell quantifies the win-rate for a challenger model outperforming an opponent model in defining the top-50 most frequently observed ICD-10-CM codes. As tie outcomes are permitted, the sum of win rates for any given model pair may total less than 100%, and a win rate of 0.5 should not be interpreted as random chance performance.

Based on the constructed win-rate matrix, we applied ILSR to compute the maximum-likelihood estimate of the stationary distribution $\pi$ for the five candidate models ["PubMedGPT2", "Llama2", "Mistral", "MedLlama2", "BioMistral"]. The resulting condition logits were estimated as [-2.067, -0.059, 0.376, 0.541, 1.208], which yield the following model selection probabilities: [0.017, 0.124, 0.192, 0.226, 0.441] based on (3). The estimation suggests that BioMistral has the highest probability of being selected as the optimal base model for downstream ICD-10 coding fine-tuning.

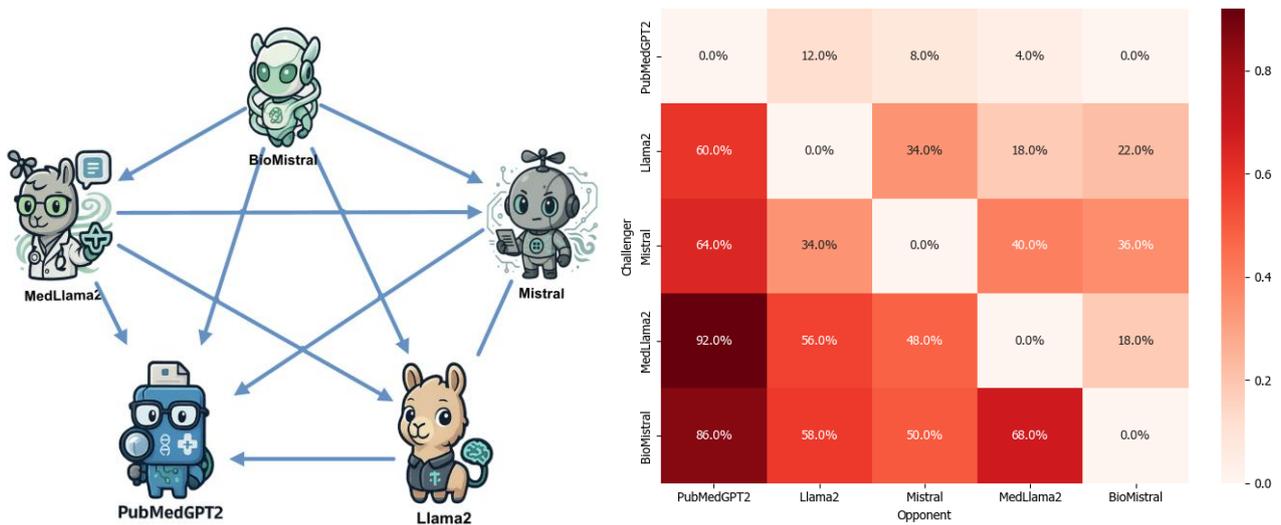

**Fig. 5.** Results of the LLM-as-judge evaluation for intrinsic ICD-10 comprehension. The left panel illustrates pairwise model comparisons using a mixed graph, where each node represents a base model. The right panel shows a win-rate matrix heatmap, where each cell quantifies the proportion of matchups in which a challenger model (row) outperformed an opponent model (column) in generating definitions for the top 50 most frequent ICD-10-CM codes.

*4.2.2 Effectiveness of base model selection*

We discuss the effectiveness of the proposed base model selection strategy from two complementary perspectives: (1) the appropriateness of using the LLM-as-judge framework for intrinsic model evaluation, and (2) the utility of the Plackett-Luce model, via ILSR, for aggregating pairwise results into a coherent global ranking.

First, using the LLM-as-judge paradigm for pairwise model evaluation proved effective in identifying models with strong intrinsic comprehension of ICD-10-CM codes. Matchup results from our framework align with prior findings reported by Lee and Lindsey [4], who evaluated LLMs using word-overlap metrics between LLM-generated outputs and ground-truth code definitions. Specifically, as visualized in the mixed comparison graph in Fig. 5, the relationship MedLlama2 > Llama2 mirrors their reported trends, suggesting that domain-adapted models possess stronger semantic understanding for diagnostic terminology. Notably, in our experiment, BioMistral consistently achieved the highest win rates (50%–86%) across all pairwise comparisons, establishing itself as the most competent model for ICD-10-CM definition generation. MedLlama2 also demonstrated competitive performance, with win rates ranging from 48% to 92%. In contrast, PubMedGPT2, despite being fine-tuned on 500k PubMed abstracts, underperforms relative to general domain LLMs such as Llama2 and Mistral. Its limited parameter size (774M) may have caused this, which restricts representational capacity and leads to comparatively weaker win rates across all matchups.

Interestingly, our results diverge from those of Lee and Lindsey [4], who reported that Mistral outperformed Llama2 and MedLlama2 in broader coding tasks. We attribute this discrepancy to fundamental differences in task scope and formulation. Their benchmark encompassed a diverse array of medical coding systems, ranging from ICD-9/10 to current procedural terminology codes, national drug codes, and logical observation identifiers, names and codes. It employed a multi-code generation setting that required models to output multiple code definitions in a single response. Such a design introduces added complexity related to instruction following and response formatting, which may confound the assessment of raw code comprehension.

By contrast, our framework isolates semantic understanding through a single-code definition task, allowing the model to generate one ICD-10-CM code per prompt. This targeted formulation minimizes confounding variables and centres the evaluation on intrinsic ICD-10-CM understanding. By isolating the generation of individual definitions, our task design emphasizes intrinsic ICD-10-CM comprehension, allowing for a more interpretable and focused evaluation of a model's diagnostic coding knowledge. These differences in task design likely account for the observed divergence in model rankings between the two studies. Furthermore, our method relies on pairwise comparative assessments rather than global quality scores, a strategy endorsed by Levtsov and Ustalov [43], who argue that pairwise evaluation offers a more nuanced and interpretable metric for open-ended or subjectively judged outputs, which is better suited for tasks where defining a single comprehensive quality metric, such as ICD code comprehension, is inherently difficult.

Second, we discuss the model ranking via the Plackett-Luce aggregation method. While one might consider simply selecting the model with the highest cumulative win-rate across all pairwise comparisons, as shown in Fig. 5, such an approach quickly becomes impractical when the number of candidate models increases or certain model matchups are computationally expensive or infeasible. We apply the Plackett-Luce model, implemented through ILSR, to aggregate available pairwise results into a stable and interpretable ranking to address these limitations.

This approach proves particularly advantageous in scenarios with partial comparison matrices, such as when a new model is introduced late in the evaluation pipeline. To demonstrate this, we manually modified the full win-rate matrix in Fig. 5 to create two hypothetical scenarios where BioMistral only competes against (1) PubMedGPT-2 and Llama2 (Fig. D.1), and (2) PubMedGPT-2 and Mistral (Fig. D.2). Despite this limited pairing, the proposed method successfully infers a complete global ranking, estimating conditional logits and selection probabilities for all five models similar to the full comparison matrices. For instance, the estimated selection probabilities in Fig. D.1 were [0.017, 0.125, 0.169, 0.276, 0.412] for PubMedGPT-2, Llama2, Mistral, MedLlama2, and BioMistral, respectively, with a mean squared error less than 0.0008. This demonstrates the flexibility and robustness of the proposed method in estimating model quality even when only partial matchups are available. Such a capability is essential for scalable model benchmarking, especially as the landscape of foundation models continues to evolve rapidly.

4.3. Experiment 2: ICD-10 coding task performance comparison

*4.3.1 Comparative visualization of model capability and performance*

Fig. 6 illustrates a comparative analysis of five developed decoder-based models, visualized in a two-dimensional space defined by win-rate (x-axis) and F1-score (y-axis). The bubble size corresponds to the parameter count of each model. This visualization reveals a clear positive correlation between the model's intrinsic ICD-10 comprehension capability and downstream ICD-10 coding performance on the internal test set, suggesting that fine-tuned models with stronger ICD-10 coding performance also tend to outperform others in pairwise semantic comprehension tasks.

**Fig. 6.** Characteristics of the fine-tuned model characteristics across average win-rate (x-axis), F1-scores (y-axis), and model parameter size (bubble size).

Among the evaluated models, BioMistral achieves the highest overall performance with an average win-rate of 0.692 and an F1-score of 0.78, positioning it in the upper-right quadrant of the performance space. MedLlama2 and Mistral exhibit comparable performance, with approximately 0.535 and 0.435 win rates, respectively. Despite sharing a similar parameter size of 7B, Llama2 demonstrates lower

PR-scores, resulting in a moderate F-score of 0.749 with a win-rate of 0.335. By contrast, PubMedGPT2, despite being pretrained explicitly on biomedical abstracts, demonstrates the weakest performance across all metrics. Its relatively small parameter size of 774M results in the lowest F1-score (0.663) and win-rate (0.06). The visualization results align closely with the rankings suggested by our base model selection framework, validating its effectiveness in guiding optimal model selection for downstream ICD-10-CM coding tasks.

*4.3.2  Fine-tuned model evaluation with the DischgDiag section*

Table 4 further presents the detailed evaluation results for all developed models on the internal test set, where we only used the DischgDiag section as input. Overall, decoder-based models consistently outperform other approaches across all metrics. The fine-tuned BioMistral achieves the highest F1-score, reinforcing its intrinsic and extrinsic top performance both intrinsically and extrinsically. Non-decoder models, in contrast, tend to struggle with R-scores. While some may achieve high P-scores, their inability to comprehensively identify relevant ICD-10 codes limits their performance. For example, both BERT and BiGRU achieve higher P-scores among non-decoder and decoder-based models; their inability to cover the full set of relevant ICD-10 codes leads to diminished effectiveness for comprehensive medical coding tasks. These findings underscore the advantage of decoder-based architectures, particularly those guided by principled model selection and fine-tuning strategies for ICD-10-CM coding applications.

**Table 4**
ICD-10-CM coding task performance comparison.

Prior literature [3] has reported suboptimal performance of decoder-based models such as GPT-4, Gemini Pro, and Llama2-70b on medical code querying without task-specific fine-tuning. Their performance for the ICD-10 coding task lagged behind much smaller, pipeline-based BERT-driven systems [44]. Our previous study [7] demonstrated the feasibility of applying a decoder-based model for ICD-10 coding tasks. The current results, as shown in Table 4 and visualized in Fig. 6, further validate the efficacy of our base model selection strategy, using intrinsic evaluations and the LLM-as-judge framework to reliably identify and prioritize models that are most amenable to downstream fine-tuning for ICD-10-CM tasks, reinforcing the practical value of adopting principled selection mechanisms in clinical NLP pipelines, especially amid the rapid evolution of foundation models.

4.4. Experiment 3: Effectiveness of redundancy-aware sampling and section-wise content inclusion

To access the effectiveness of the proposed redundancy-aware sampling strategy and to evaluate the impact of including section-wise content on ICD-10 coding performance, we conducted further analysis using BioMistral, the top-ranked model identified by our base model selection framework, in the following subsections.

*4.4.1 Performance gains from redundancy-aware sampling*

We evaluated two BioMistral configurations trained exclusively on the DischgDiag section: one on the full dataset and the other on a deduplicated dataset constructed via our redundancy-aware sampling procedure. As shown in Fig. 7, the deduplicated model consistently outperformed the baseline across PRF-scores, under the full test set and a targeted subset of test cases containing all five considered sections (corresponding to the final row of Table 3). Note that this additional subset evaluation allows for a direct performance comparison between the DischgDiag-only model and the universal model trained with multi-section inputs (evaluated in Sections 4.4.2 and 4.4.3), thereby isolating the impact of the proposed redundancy-aware sampling procedure independent of input structure complexity.

**Fig. 7.** Comparison of the ICD-10-CM coding performance for the BioMistral model trained on the full dataset versus the deduplicated dataset. Results are reported for both the full test set (left panel) and a subset of test cases containing all section combinations (right panel).

Beyond performance gains, the deduplicated model also achieved a 10.2% reduction in total training time, highlighting the computational efficiency and resource savings enabled by removing redundant samples. These findings align with prior studies [29, 30, 45], which emphasizes that deduplication not only enhances training efficiency and reduces computational costs but also enhance model generalization and accuracy on downstream tasks especially in domains like clinical documentation where repetitive phrasing is prevalent.

*4.4.2 Impact of section-wise content inclusion on model performance*

This subsection investigates how incorporating additional structured sections beyond DischgDiag affects ICD-10-CM coding performance. Starting with DischgDiag as the base input, we incrementally added OpNote, MedHist, PathRep, and TreatCous following the section priority specified in Table 3, resulting in five section combination-specific models. We evaluated each model under two conditions: 1) Full test set: we included all test samples, regardless of their section composition. 2) Matched test subset: Where we only included test samples containing the exact section combination used during model training (as specified in the "Reports Containing Combination (Test)" column of Table 3). Table 5 presents the comparative results.

**Table 5**

Overall, we observe a consistent trend of improved F-scores as additional structured sections are incorporated, but only when we evaluate the models on their matched section-combined subsets. For instance, the model trained with DischgDuag+OpNote+MedHist+PathRep achieves a 0.011 F-score gain over the DischgDiag-only model on the matched subset, suggesting that including more clinically relevant sections improves the model's capacity to capture diagnostic cues when aligned with appropriately structured inputs.

Among the individual sections, MedHist contributed the most significant performance gains. In an additional comparison, the DischgDiag+MedHist model elevated the PRF-scores from 0.798/0.776/0.787 to 0.803/0.786/0.794 on the matched subset, highlighting the diagnostic value of medical history content. In contrast, including OpNote alone resulted in a marginal drop in overall model performance, as shown in Table 5, suggesting limited utility for ICD-10-CM code prediction. A follow-up consultation with CCSs confirmed that OpNotes primarily support ICD-10 procedure codes and offer relatively little utility for diagnostic code assignment.

Nonetheless, when we combined OpNote with MedHist, we observed a slight gain in P-score (+0.004) and a modest increase in F-score by +0.002 relative to including MedHist alone. Although OpNote may not be impactful in isolation, it provides complementary contextual value when paired with other diagnostically relevant sections. These findings align with clinical documentation practices: OpNotes often reinforce or confirm diagnoses already established elsewhere in the record, thereby supporting code refinement rather than expansion. Their inclusion can help suppress irrelevant or uncertain codes, improving coding specificity. Therefore, OpNote may remain in model inputs, particularly when input length constraints allow for potential precision gains in ICD-10-CM coding.

Importantly, this performance improvement does not consistently extend to the full test set, which includes a heterogeneous mix of section configurations. While the five-section model maintains a modest F-score improvement of +0.011 over the DischgDiag-only baseline on the full test set, models trained on other partial section combinations underperform when evaluated outside their matched section structure. This performance drop highlights a core limitation of section-combination-specific models. While effective under structurally aligned conditions, their generalizability to discharge summaries with variable or incomplete section content is reduced. These observations underscore the need for models that are not only section-aware but also robust to real-world variability in clinical note structure. To address this challenge, we explore the development of a universal model in Section 4.4.3, capable of handling diverse section combinations using a unified training and inference strategy.

*4.4.3 Universal model performance*

To evaluate the generalizability and robustness of a single unified model, we compared the universal model's F-scores based on variable section combination inputs with those of the section-combination-specific models presented in Table 5. The comparative results are illustrated in Fig. 8. At the same time, detailed metrics including PRF-scores and MDCA are provided in Table C.1. Unlike the section-combination-specific approach, which trains and applies a separate model for each distinct input structure, the universal model uses a single architecture trained to accommodate and adapt to varying section combinations through a unified prompt format.

Notably, the universal model consistently demonstrates performance improvements as more structured sections are included in the input, both on the full test set and the matched section-combined subsets. This contrasts with the behavior observed in section-specific models, which generally perform well only under structurally aligned conditions. In particular, the R score increased by +0.3 as more sections were included, reflecting the model's enhanced ability to identify more relevant ICD-10-CM codes when additional context is available. This gain is clinically significant, as higher R reduces the risk of under-coding, helping prevent potential revenue loss and ensuring more complete clinical documentation. Furthermore, comparing results across Figs. 7 and 8, and Table 5 further reinforces the synergistic effect of redundancy-aware sampling and section-wise content enrichment strategies. Specifically, the F-score improves from 0.780 (baseline), 0.787 (+Redundancy-aware sampling) to 0.802 (full section inputs) on the full test set, and from 0.773 (baseline), 0.783 (+Redundancy-aware sampling) to 0.800 (full section inputs) on the section-combined subset. These incremental improvements validate the integrated benefits of deduplication and structured input augmentation, culminating in superior performance for ICD-10-CM coding.

**Fig. 8.** Comparison of ICD-10-CM coding performance between the universal model and section-combination-specific models. The left/right panels compare the performance of the universal model against individually fine-tuned models on the full test set/section-combined-test subset. The F-scores were used as the evaluation metric.

As depicted in the right panel of Fig. 8, the universal model demonstrates competitive performance compared to section-specific models across various input configurations, especially when two or more structured sections are presented. Notably, when we included MedHist in the input, the universal model outperforms all corresponding section-specific counterparts on the full test set, highlighting its flexibility and generalize across varied documentation formats. However, in scenarios where only the DischgDiag section is presented, either due to intentional masking or genuine omission of other sections, the section-specific model trained solely on DischgDiag surpasses the universal model by an F-score margin of approximately 0.02, suggesting that section-specific fine-tuning may still be advantageous in resource-constrained or sparse input scenarios, where discharge summaries lack comprehensive section coverage or sufficient contextual information.

These results reveal an inherent trade-off between generalizability and specialization. While the universal model performs robustly across diverse input structures, section-specific models remain preferable when input content is limited. Ultimately, the universal model is most effective when discharge summaries include multiple high-value sections, enabling it to fully leverage structured richer context for more accurate ICD-10-CM code prediction.

4.5. Experiment 4: Cross-institutional robustness and generalizability of ICD-10-CM coding models

Table 6 shows the external validation results using datasets collected from two independent hospitals. The upper part of the table corresponds to data curated by CCSs at KMUH, collected during routine ICD-10-CM coding workflows after we integrate the system into hospital operations. The lower part reports results on the retrospective external dataset collected from TMMH, allowing us to evaluate the generalizability of the proposed pipeline in a different institutional context without retraining.

The extended validation at the source hospital (KMUH) reveals trends consistent with those observed on the internal test set (Table 4). Specifically, non-decoder-based models such as HAN and BiGRU exhibit low R-scores, severely limiting their overall performance. Although BERT achieves the highest P-score again, its poor R-score results in a modest F-score of 0.421. In contrast, decoder-based LLMs consistently outperform traditional models. Aligned with the findings from our model selection framework, BioMistral achieves the best overall performance, with an F-score of 0.754 on full code prediction and 0.878 on the top-50 most frequent codes. While MedLlama2 delivers comparable results and identical MDCA, BioMistral remains the top performer across all evaluation metrics.

Furthermore, we observe performance gains when applying redundancy-aware sampling, which improves BioMistral's F1-score to 0.764 and raises MDCA to 0.627. Incorporating all five structured sections during inference using the universal model yields the best overall performance, achieving an F-score of 0.779 on full codes, 0.920 on top-50 codes, and the highest MDCA of 0.728. These findings confirm the effectiveness of the proposed pipeline, including base model selection, data refinement, and context-enriched input strategies for enhancing model performance in real-world hospital workflows.

**Table 6**

ICD-10-CM coding task performance on the external datasets.

As expected, the performance of all models declined when evaluated on the external TMMH dataset. Traditional models performed poorly; for example, all non-decoder-based models recorded R-scores below 0.2, and BERT's P-score dropped sharply from 0.875 (at the extended KMUH dataset) to 0.137. Likewise, PubMedGPT-2 demonstrated weak generalizability, with its F-score falling below 0.4. Nevertheless, decoder-based LLMs with 7B parameters, especially BioMistral, again showed strong relative performance. The redundancy-aware version of BioMistral yielded a 0.626 F-score on the full code set, representing a +0.06 improvement over its non-deduplicated counterpart. The universal model incorporating all available sections as its inputs achieved the best results on the TMMH dataset, with an F-score of 0.636, a top-50 F-score of 0.757, and an MDCA of 0.573.

These findings demonstrate that the proposed pipeline performs strongly within the source institution and generalizes effectively to external hospital data. The synergistic integration of redundancy-aware sampling and structured section enrichment significantly enhances model robustness and accuracy, underscoring the pipeline's suitability for deployment across diverse clinical settings.

## 5. Limitations

While this study provides valuable insights into the performance of LLMs for ICD-10-CM code extraction from discharge summaries, we acknowledge several limitations. First, our work focuses exclusively on ICD-10-CM diagnostic codes and does not address procedure coding. Given the clinical and financial importance of the procedure codes, future work should explore extending the proposed pipeline to support multi-type medical coding tasks. Second, the study only investigates ICD-10-CM code extraction from discharge summaries of hospitalized patients. The results may not generalize to other clinical document types, such as outpatient visit notes or daily inpatient progress notes, which could be structurally and semantically different from discharge summaries.

All ICD-10-CM codes in the dataset reflect the CCSs' coding decisions during specific periods. However, diagnostic coding practices evolve due to changes in clinical guidelines, institutional policies, coder training, and billing optimization strategies. As a result, unless we perform periodic retraining or domain adaptation, the trained models may be subject to coding drift, potentially degrading performance in longitudinal deployments.

The primary training dataset was sourced from a single institution (KMUH) and spans a relatively narrow timeframe (April 2019 to March 2021). Although we conducted additional validation on later data from the same institution and one other hospital (TMMH), we collected all the data within Taiwan and from a limited number of institutions. As such, the generalizability of our models to other geographic regions or healthcare systems with different documentation styles remains unverified.

Our evaluation relied on standardized prompts for both model selection and code generation. However, prior research has shown that LLM outputs are highly sensitive to prompt wording. Using alternative prompts could yield different results, introducing variability and potential bias in performance assessments. Furthermore, our approach assumes the presence of well-structured clinical documentation, including key sections such as DischgDiag. While our universal model attempts to accommodate section variability, performance may decline when summaries lack key sections or input exceeds token limits and requires truncation.

Finally, modifications to prompt templates, documentation style shifts, or clinical coding guidelines updates may necessitate model retraining, a resource-intensive and time-consuming process, underscoring a broader limitation of static, model-based approaches in adapting to evolving knowledge systems. Future work should consider alternative paradigms such as retrieval-augmented generation or federated learning, which may offer greater flexibility in handling knowledge drift and institutional heterogeneity without requiring extensive retraining.

## 6. Conclusion

This study presents a modular and principled framework for ICD-10-CM code prediction using open-source LLMs, addressing critical challenges in base model selection, data efficiency, and contextual input modeling. We introduced an LLM-as-judge-based evaluation framework, combined with Plackett-Luce aggregation, to assess and rank candidate base models based on their intrinsic understanding of ICD-10-CM code definitions prior to task-specific fine-tuning. This method offers a scalable and interpretable solution to base model selection, particularly in scenarios involving numerous pretrained models.

To improve training efficiency and data utility, we proposed a redundancy-aware sampling strategy that eliminates semantically duplicated samples based on embedding similarity and code overlap. Our results show that this approach reduces dataset size and improves downstream performance across multiple evaluation metrics. Leveraging the structured format of Taiwanese electronic medical records, we systematically analysed the impact of section-wise content inclusion, providing a foundation for future integration with HL7/FHIR standards. Our findings demonstrate that integrating high-value sections (e.g., DischgDiag and MedHist) consistently improves ICD-10-CM coding accuracy. Moreover, we evaluated universal and section-specific modelling strategies, revealing a trade-off between generalizability and specialisation: universal models perform robustly across variable inputs, while section-specific models retain advantages in sparse or incomplete documentation scenarios. Extensive experiments conducted across two hospitals in Taiwan validate the robustness and portability of the proposed pipeline, demonstrating strong performance even without retraining on external institutional data.

These contributions underscore the feasibility of deploying open-source LLMs for real-world clinical coding tasks. Our framework provides a practical, scalable solution that supports high-quality, automated ICD-10-CM coding, offering actionable insights for future implementations of LLM-driven coding assistance systems in diverse healthcare settings.

## Acknowledgements

This research was supported by funding from the project "AI-assisted ICD-10-CM/PCS coding model construction to improve coding quality and operational performance" funded by Kaohsiung Medical University Chung-Ho Memorial Hospital and Taitung MacKay Memorial Hospital under the project number TMMH-114-05, as well as by the National Science and Technology Council under grant number NSTC 112-2221-E-992 -056-MY3. The authors would like to thank the participating hospitals for providing access to anonymized clinical data and CCSs for their valuable insights during system development and evaluation. We also acknowledge the contributions of our collaborators and reviewers for their constructive feedback throughout the research process.

## Funding sources

Funding: This work was supported by Kaohsiung Medical University Chung-Ho Memorial Hospital in the "AI-assisted ICD-10-CM/PCS coding model construction to improve coding quality and operational performance" project; Taitung MacKay Memorial Hospital [grand number TMMH-114-05]; and the National Science and Technology Council [grand number NSTC 112-2221-E-992 -056-MY3].